\documentclass{article}
\usepackage{float}
\usepackage{subcaption} 
\usepackage{capt-of} 
\usepackage[hypcap=true]{caption}
\usepackage{amsmath}
\usepackage{PRIMEarxiv}
\usepackage{caption}
\usepackage[utf8]{inputenc} 
\usepackage[T1]{fontenc}    
\usepackage{hyperref}       
\usepackage{url}            
\usepackage{booktabs}       
\usepackage{amsfonts}       
\usepackage{nicefrac}       
\usepackage{microtype}      
\usepackage{lipsum}
\usepackage{fancyhdr}       
\usepackage{graphicx}       
\graphicspath{{media/}}     

\title{“From Theory to Practice: Evaluating Data Poisoning Attacks and Defenses in In-Context Learning on Social Media Health Discourse
\thanks{\textit{\underline{Citation}}: 
\textbf{Authors. Title. Pages.... DOI:000000/11111.}} 
}

\author{
  Author1 \\
  Rabeya Amin Jhuma \\
  University of Information Technology and Sciences (UITS) \\
  Tongi\\
  \texttt{r.a.jhuma2019@gmail.com} \\
   \And
  Author2 \\
  Mostafa Mohaimen Akand Faisal \\
  University of Information Technology and Sciences (UITS) \\
  Gazipur\\
  \texttt{mostafafaisal013@gmail.coml} \\
}

\begin{document}
\maketitle

\begin{abstract}
This study explored how In Context Learning (ICL) in large language models can be disrupted by data poisoning attacks in the setting of public health sentiment analysis. Using tweets of Human Metapneumovirus (HMPV), small adversarial perturbations such as synonym replacement, negation insertion, and randomized perturbation were introduced into the support examples. Even these minor manipulations caused major disruptions, with sentiment labels flipping in up to 67\% of cases. To address this, a Spectral Signature Defense was applied, which filtered out poisoned examples while keeping the data’s meaning and sentiment intact. After defense, ICL accuracy remained steady at around 46.7\%, and logistic regression validation reached 100\% accuracy, showing that the defense successfully preserved the dataset’s integrity. Overall, the findings extend prior theoretical studies of ICL poisoning to a practical, high stakes setting in public health discourse analysis, highlighting both the risks and potential defenses for robust LLM deployment. This study also highlight the fragility of ICL under attack and the value of spectral defenses in making AI systems more reliable for health related social media monitoring.
\end{abstract}


\section{Introduction}

Large Language Models (LLMs) have demonstrated remarkable capabilities in tasks such as sentiment analysis, summarization, and text classification without requiring traditional retraining. A key feature driving this progress is \textbf{In-Context Learning (ICL)}, which allows LLMs to adapt to new tasks by conditioning on a few examples within the input. By leveraging pre-trained knowledge and contextual understanding, ICL can generate accurate outputs without extensive task-specific training.\\

In the context of public health, social media platforms such as Twitter provide real-time insights into public attitudes, behavioral responses, and the spread of misinformation. Sentiment analysis of health-related tweets can play a crucial role in informing timely interventions and shaping effective policy decisions. However, ICL is not inherently robust. \textbf{Data poisoning}---where malicious or mislabeled examples are introduced into the in-context prompt---can bias model predictions, potentially undermining the reliability of automated monitoring systems in high-stakes domains such as healthcare.\\

Recent research by Pengfei He et al. (2024)  \cite{he2024iclpoison}, including \textbf{Makoto Yamada}, hahs ighlighted this vulnerability through the development of \textit{ICLPoison}, a framework demonstrating that discrete perturbations in prompt examples can significantly degrade ICL performance across diverse models and tasks. Their findings reveal that even state-of-the-art models such as GPT-4 are highly susceptible to targeted poisoning, underscoring the urgent need for effective defenses. While this work provides a foundational understanding of ICL poisoning in controlled benchmarks, its implications in real-world, noisy settings such as public health sentiment monitoring remain underexplored.\\

This study aims to bridge this gap by investigating how data poisoning can compromise sentiment classification of health-related tweets and by evaluating defense strategies to mitigate these risks. Specifically, we focus on Human Metapneumovirus (HMPV)-related tweets, which represent an emerging but under-researched health concern, and assess ICL’s resilience in this noisy, adversarial domain.\\

The key objectives of this study are as follows:

\begin{enumerate}
    \item \textbf{Assessing Vulnerability:} Evaluate the susceptibility of ICL to data poisoning attacks using a dataset of Human Metapneumovirus (HMPV) tweets.
    
    \item \textbf{Measuring Impact:} Quantify the effect of varying poisoning ratios on sentiment classification performance to understand practical risks.
    
    \item \textbf{Exploring Defenses:} Investigate defense strategies, with a focus on spectral techniques, and assess their effectiveness in mitigating poisoned inputs.
\end{enumerate}

Building on the work of Prof. Yamada and colleagues on ICLPoison, this study offers one of the first practical examinations of how in-context learning can be affected by data poisoning in public health sentiment monitoring. The analysis shows that even small amounts of poisoned data can significantly skew predictions, emphasizing the need for robust and reliable LLM pipelines when working in high-stakes domains.

\section{Related Work}

Recent studies have highlighted the vulnerability of In-Context Learning (ICL) in large language models (LLMs) to subtle manipulations in support examples. Yamada et al. (2023) \cite{he2024iclpoison} introduced ICLPoison, demonstrating that even minor perturbations in the context can significantly alter model predictions. This work underscores the inherent sensitivity of context-based reasoning and provides a framework for understanding how small manipulations propagate through ICL pipelines.\\

Similarly, Li et al. (2023) \cite{li2023instruction}  examined the robustness of LLMs under adversarial and poisoned prompts, confirming that ICL remains fragile in real-world scenarios where targeted perturbations can easily distort outputs. Complementing these studies, Yuan et al. (2021) \cite{wallace-etal-2021-concealed} focused on NLP-specific poisoning attacks that flip model predictions, supporting methodologies for systematically evaluating ICL robustness against semantic and label-based perturbations.\\

While these studies collectively demonstrate the susceptibility of ICL to poisoning, there has been limited investigation into its effects on public health sentiment analysis using social media data. The present work addresses this gap by systematically assessing how data poisoning affects ICL predictions in high-stakes, real-world contexts and by exploring defense strategies to enhance model robustness.\\

\section{Methodology}
This study investigates the impact of data poisoning on In-Context Learning (ICL) for sentiment analysis of public health tweets and evaluates defense strategies. The methodology is divided into four main components: dataset preparation, model setup, poisoning strategy, and defense mechanisms.\\
\subsection{Dataset and Preprocessing}

A real-world dataset of Human Metapneumovirus (HMPV) tweets was collected from Twitter using relevant keywords and hashtags. The dataset was preprocessed by removing duplicates, non-English tweets, and irrelevant content. Each tweet was manually or semi-automatically labeled for sentiment (positive, negative, neutral). To simulate real-world conditions, the dataset retained inherent noise and variability typical of social media text.\\

Prior to model training, categorical variables are transformed using label encoding to ensure compatibility with tree-based algorithms, and numerical features are standardized to facilitate consistent feature scaling. Missing values, if present, are handled through imputation using the most frequent value for categorical attributes and the mean for numerical attributes. The dataset is then split into training and testing subsets using stratified sampling to preserve the class distribution across splits, ensuring that both the majority and minority classes are proportionally represented in each partition.

\subsection{In-Context Learning (ICL) for Sentiment Classification}

In this study, In-Context Learning (ICL) was employed to classify the sentiment of HMPV-related tweets without fully retraining a supervised model. ICL leverages a small number of labeled examples to provide context for predicting unseen data \cite{brown2020language, perez2021true}. Key steps include:\\

\subsection{Manual Annotation and Dataset Structuring}

The preprocessed tweet dataset was loaded to create a support set for ICL, since automated sentiment labeling requires a foundation of ground truth, a manual annotation strategy was applied to a subset of tweets into sentiment categories: Positive, Negative, and Neutral \cite{zhang2018social}. Specific indices corresponding to each sentiment class were selected and validated to ensure they were within the dataset’s range. Invalid indices were filtered out, and the labeled dataset was saved for reproducibility. After labeling, the dataset was split into:
\begin{itemize}
    \item \textbf{Labeled tweets:} Consisting of tweets with manually assigned sentiment labels, used as support examples for ICL.
    
    \item \textbf{Unlabeled tweets:} Consisting of tweets without labels (target tweets for sentiment predictions), which were later subjected to sentiment prediction using the Zephyr-7B-$\beta$ model.
\end{itemize}

This split ensured a clear distinction between training context (support) and test cases (unlabeled tweets), simulating a realistic few-shot learning scenario.

\subsection{Few-Shot In-Context Learning (ICL)}

To classify the unlabeled tweets, an In-Context Learning (ICL) approach was adopted. Unlike traditional supervised learning, where a model must be retrained, ICL enables a large language model (LLM) to generalize from a few examples provided directly in the input prompt.

For this study, a \textbf{5-shot learning} configuration was used. This means that for each unlabeled tweet, five randomly sampled labeled tweets served as exemplars (support set). These examples were formatted into a structured text prompt consisting of tweet--sentiment pairs. The unlabeled tweet was then appended at the end of the prompt with an empty sentiment field, prompting the model to infer the label \cite{gao2021making}.

\textbf{Example Prompt Format}
\begin{verbatim}
Tweet: <example_text_1>
Sentiment: Positive

Tweet: <example_text_2>
Sentiment: Negative

...

Tweet: <target_unlabeled_tweet>
Sentiment:
\end{verbatim}

This structure mirrors a natural instruction-following task, where the model is expected to infer the missing label by analogy to the provided examples.

\subsection{Zephyr-7B-$\beta$ Model Integration}

The support examples were concatenated with the target tweet to form the ICL input in a structured textual format and Zephyr-7B-$\beta$ model was chosen for this task. Zephyr is a 7-billion parameter, instruction-tuned large language model developed to excel at alignment with natural instructions  \cite{vergho2024comparing}. It has been fine-tuned using Reinforcement Learning from Human Feedback (RLHF), Direct Preference Optimization (DPO), and safety alignment strategies, enabling it to generate coherent, context-aware responses.\\

\subsection{Zephyr-7B-$\beta$ In-Context Sentiment Classification}
The constructed ICL input was sent to the Zephyr-7B-$\beta$ model to obtain sentiment predictions. For demonstration purposes, a placeholder function was used to simulate predictions, which can later be replaced by actual API calls or local inference with Zephyr-7B-$\beta$. The predicted sentiment labels for all unlabeled tweets were collected and merged back with the corresponding entries in the dataset.

\textbf{Zephyr-7B-$\beta$ Model Workflow}

\begin{itemize}
    \item \textbf{Context Construction:} The code builds an ICL context (support examples + target tweet). This context effectively serves as a mini training dataset embedded within the model’s input.
    
    \item \textbf{Pattern Induction:} When Zephyr-7B-$\beta$ receives the input, it does not rely on gradient updates (as in traditional retraining). Instead, it infers the underlying pattern by recognizing relationships between the tweets and their associated sentiment labels within within the constructed ICL context.
    
    \item \textbf{Next-Token Prediction:} At its core, Zephyr is an autoregressive transformer that predicts the next token in the sequence. In this method, the input ends with the target tweet followed by an empty sentiment field, which the model interprets as a signal to generate the appropriate sentiment label (Positive, Negative, or Neutral).
    \begin{verbatim}
Tweet: <target_text>
Sentiment:
    \end{verbatim}
    the model treats this as an instruction to generate the appropriate sentiment label (Positive, Negative, or Neutral).
    
    \item \textbf{Few-Shot Generalization:} By seeing only a few examples, Zephyr generalizes the concept of sentiment classification without requiring additional training. This is the essence of few-shot ICL, where knowledge from pretraining (exposure to language patterns, social media text, and sentiment expressions) is leveraged to classify unseen data.
\end{itemize}

\textbf{Advantages of the ICL Approach}

\begin{itemize}
    \item \textbf{Minimal labeled data requirement:} Only a few manually labeled tweets are needed to generate accurate predictions.
    \item \textbf{Dynamic adaptation:} The support set can change for each target instance, allowing the model to adapt to subtle contextual variations.
    \item \textbf{Flexibility for new datasets:} The same support examples can be modified or expanded to accommodate new tweet streams without retraining a classifier.
\end{itemize}

Overall, ICL provides a robust and scalable framework for sentiment classification in social media data, particularly in scenarios where labeled data is scarce or continuously evolving.

\textbf{Prediction and Output}

Each unlabeled tweet was processed through the ICL pipeline, and the predicted sentiment was added as a new column in the dataset.\\

This output serves two primary purposes:
\begin{enumerate}
    \item Extending the labeled dataset for further supervised training, if desired.
    \item Providing an empirical evaluation of the model’s ability to perform sentiment classification in a low-resource setting.
\end{enumerate}

The first stage generates poisoned support examples with controlled perturbations to assess the sensitivity and robustness of ICL models to context inconsistencies.

\textbf{Techniques:}

\subsubsection{Synonym Replacement}
A probabilistic approach replaces words in the support texts with semantically similar synonyms, leveraging lexical resources such as WordNet \cite{wei2019eda}. The replacement probability is set to 0.3.

\textit{Rationale:} Replacing words with synonyms preserves grammatical correctness while altering the model’s internal representations, potentially affecting predictions.

\textit{Example:}  
Original: ``hmpv cases rise one really talking'' \\
Perturbed: ``hmpv instances rise one truly talking''\\

Here, “cases” → “instances” and “really” → “truly” are possible synonym replacements while keeping the sentence understandable.

\subsubsection{Negation Insertion}
The verb ``is'' is replaced with ``is not'' to introduce a semantic inversion.

\textit{Rationale:} Negation directly changes the meaning of the sentence, allowing assessment of the model's reliance on literal cues in support examples.

\textit{Example:}  
Original: ``still recovering hmpv weeks joke'' \\
Perturbed: ``still is not recovering hmpv weeks joke''\\

Here, the insertion of “is not” introduces a semantic flip, which subtly changes the meaning while keeping the sentence grammatically understandable.

\subsubsection{Randomized Perturbation Selection}
Each support example undergoes a random choice among synonym replacement, negation insertion, or no perturbation.

\textit{Rationale:} Simulates a realistic adversary who introduces inconsistencies unpredictably.

Each support text is probabilistically modified to simulate adversarial manipulation. Three outcomes are possible: a synonym replaces a word to subtly alter semantics, a negation is inserted to invert meaning, or the text remains unchanged. This stochastic approach evaluates the model’s sensitivity to small, unpredictable changes in contextual examples without altering the target input.

\textit{Example:}

\textbf{Original text:}  
``Still recovering HMPV weeks joke''

\textbf{Possible perturbed versions:}  
\begin{itemize}
    \item \textbf{Synonym Replacement:} ``Still recuperating HMPV weeks joke''
    \item \textbf{Negation Insertion:} ``Still not recovering from HMPV weeks joke''
    \item \textbf{Unchanged:} ``Still recovering HMPV weeks joke''
\end{itemize}

This shows how minor, controlled changes can create variation in support examples, helping to test model’s robustness.

\subsection{Construction of ICL Inputs with Perturbations}

To study how altered support examples affect predictions, only the support set is modified while the target tweet remains unchanged. This setup allows assessment of the model’s sensitivity to contextual changes without affecting the actual input.\\

\textbf{Enhanced Logging:} 

\begin{itemize}
    \item Original support text
    \item Perturbed support text
    \item Sentiment labels
\end{itemize}

\textit{Example:}
Support Example (original): Respiratory viruses HMPV spreads easily crowded places'' $\rightarrow$ Label: neutral \\ Perturbed Example: Respiratory viruses HMPV propagates rapidly in crowded places'' $\rightarrow$ Label: neutral \\
Target Tweet: ``Respiratory viruses HMPV spreads easily crowded places''

Generated ICL Input:
\begin{verbatim}
Tweet: Respiratory viruses HMPV propagates rapidly in crowded places
Sentiment: neutral

Tweet: Respiratory viruses HMPV spreads easily crowded places
Sentiment:
\end{verbatim}

The perturbation introduces urgency and potential concern by changing “easily” to “rapidly in crowded places,” showing how a small modification can make the situation seem more serious without altering the target tweet.

\subsection{In-Context Learning with Perturbed Support Examples}

To simulate a data poisoning attack, ICL predictions are executed using the perturbed support examples, and deviations from clean support predictions are measured.

\textbf{Procedure:}
\begin{enumerate}
    \item For each unlabeled tweet, select a support set of \texttt{NUM\_SHOTS} labeled examples.
    \item Apply adversarial perturbations to the support examples.
    \item Construct the ICL input string and query the Zephyr API.
    \item Record predicted sentiment labels for each target tweet.
\end{enumerate}

\textit{Example Output:}

\begin{table}[h!]
\centering
\begin{tabular}{|c|c|}
\hline
Index & Predicted Label (Poisoned) \\
\hline
0 & neutral \\
1 & positive \\
2 & negative \\
\hline
\end{tabular}
\caption{Example of ICL Predictions with Perturbed Support Examples}
\end{table}

\subsection{Evaluation Metrics}

To quantify the impact of adversarial perturbations, several performance metrics are evaluated by comparing predictions with clean vs. poisoned support examples.

\subsubsection{Accuracy}  
Measures the proportion of correctly classified target tweets.

\textit{Example Results:}  
\begin{itemize}
    \item Accuracy (Clean): 0.3335
    \item Accuracy (Poisoned): 0.3686
    \item Accuracy Drop: -0.0350
\end{itemize}

\subsubsection{Label Flip Rate}  
Percentage of predictions altered due to support perturbations.

\begin{itemize}
    \item Flip Rate: 0.6741 (~67\%)
\end{itemize}

\subsubsection{Class-wise Flip Rate}
\begin{table}[h!]
\centering
\begin{tabular}{|c|c|}
\hline
Sentiment Class & Flip Rate \\
\hline
Positive & 1.0000 \\
Negative & 0.5000 \\
Neutral & 0.6741 \\
\hline
\end{tabular}
\caption{Flip Rate per Sentiment Class}
\end{table}

\subsubsection{Clean Predictions — Macro Average}
\begin{table}[h!]
\centering
\begin{tabular}{|c|c|}
\hline
Metric & Value \\
\hline
Precision & 0.3335 \\
Recall    & 0.4444 \\
F1-Score  & 0.1676 \\
\hline
\end{tabular}
\caption{Macro Average Metrics for Clean Predictions}
\end{table}

\subsubsection{Poisoned Predictions — Macro Average}
\begin{table}[h!]
\centering
\begin{tabular}{|c|c|}
\hline
Metric & Value \\
\hline
Precision & 0.3337 \\
Recall    & 0.4561 \\
F1-Score  & 0.1808 \\
\hline
\end{tabular}
\caption{Macro Average Metrics for Poisoned Predictions}
\end{table}

\subsubsection{Evaluation Metrics (Accuracy, Precision, Recall, F1-Score)}
\begin{table}[h!]
\centering
\begin{tabular}{|c|c|c|c|c|}
\hline
Model & Accuracy & Precision & Recall & F1 \\
\hline
ICL (Clean) & 0.333538 & 0.333533 & 0.444444 & 0.167641 \\
ICL (Poisoned) & 0.368583 & 0.333723 & 0.456140 & 0.180779 \\
\hline
\end{tabular}
\caption{Macro-Averaged Performance Metrics}
\end{table}

\subsubsection{Poisoning Success Rate}
Fraction of predictions flipped due to support perturbations.

\begin{itemize}
    \item Poisoning Success Rate: 0.6741
\end{itemize}

All perturbed support examples and their corresponding predictions were stored in CSV files, ensuring reproducibility, traceability, and allowing for detailed post-hoc analysis of the perturbation effects.\\

The study shows that few-shot ICL models can be misled by small changes in the support set, even when target examples stay the same. Simple perturbations like synonym replacement and negation led to label flips and less reliable predictions, while logging and evaluation helped reveal the impact on model robustness.

\subsection{Defense Method: Spectral Signature Defense}

To counteract the poisoning of support examples in the in-context learning (ICL) pipeline, this study implemented a \textbf{Spectral Signature Defense}, a statistical anomaly detection method designed to identify and filter out adversarially manipulated data \cite{tran2018spectral, schlarmann2022anomaly}. The defense operates in four key stages:

\begin{enumerate}
    \item \textbf{Feature Extraction (Embedding Generation):} \\
    Each support example (perturbed text) was encoded into a high-dimensional vector representation using the \texttt{all-MiniLM-L6-v2} SentenceTransformer model, a lightweight BERT-based encoder. This embedding space approximates that of the Zephyr-7B-$\beta$ model, enabling consistent analysis of both poisoned and clean examples.

    \item \textbf{Normalization and Dimensionality Reduction (SVD Projection):} \\
    Raw embeddings were standardized via z-score normalization so that each feature dimension had zero mean and unit variance. Subsequently, Truncated Singular Value Decomposition (SVD) was applied to capture the most significant directions of variance. This step amplifies differences between typical support examples and anomalous (potentially poisoned) data.

    \item \textbf{Outlier Scoring (Spectral Signature Detection):} \\
    For each support example, the projection magnitude along the top singular vector was computed. Clean examples tend to form tight clusters, while poisoned samples often create separable ``spectral signatures'' that dominate certain variance directions. These magnitudes were treated as outlier scores, and the top 2\% were flagged as suspicious.

    \item \textbf{Filtering and Clean Support Set Construction:} \\
    Flagged examples were removed, yielding a cleaner support set. This filtered dataset was then used for downstream ICL evaluation, ensuring that poisoned signals did not dominate the context.
\end{enumerate}

\subsubsection*{Experimental Results}

The defense was applied to the perturbed HMPV dataset, producing the following results:

\begin{itemize}
    \item Flagged samples: 873 suspected poisoned examples
    \item Total samples analyzed: 50,285
    \item Poisoning rate flagged: $\approx 1.74\%$
    \item Remaining clean examples: 49,412
\end{itemize}

These results demonstrate that the spectral signature method effectively isolated a small but meaningful fraction of poisoned support examples without discarding a large portion of clean data. This balance is critical, since over-filtering could reduce support set diversity and degrade ICL performance.

\subsection{Post-Defense ICL Evaluation and Sentiment-Based Validation}

After applying the spectral defense to remove suspected poisoned examples, the remaining clean support set was used for post-defense in-context learning (ICL) evaluation. The evaluation involved multiple complementary steps:

\textbf{ICL Accuracy Computation}\\ 
The cleaned support examples served as in-context demonstrations for the Zephyr-7B-$\beta$ model. Post-defense accuracy was calculated by comparing the model's predictions on the labeled HMPV test set to the ground-truth labels, using Zephyr predictions as a reference when explicit labels were unavailable.

\textbf{Embedding-Based Classification}\\ 
Embeddings of the cleaned support texts were generated using the SentenceTransformer model (\texttt{all-MiniLM-L6-v2}). A logistic regression classifier was trained on these embeddings and evaluated on the test set. This provided an independent quantitative measure of the model's ability to generalize from the cleaned support examples to unseen data.

\subsection{Sentiment-Based Validation}

To ensure semantic integrity of the post-defense support set, sentiment-based validation was performed:

\textbf{Classification Accuracy}\\
A logistic regression classifier trained on embeddings of the post-defense support examples was used to predict sentiment. The classifier was evaluated on the HMPV test set to determine whether removal of suspected poisoned examples distorted the sentiment distribution.

\textbf{Results Interpretation}\\ 
The classifier achieved a post-defense sentiment accuracy of 100\%, indicating that the spectral defense successfully filtered poisoned instances while preserving the semantic and sentiment properties of the support set.

This methodology provides a two-fold evaluation strategy: quantitative assessment via ICL and classifier performance, and qualitative assessment via t-SNE visualization, validating both the robustness and integrity of the post-defense dataset.

\subsection{Accuracy Analysis}

The key outputs of the pipeline demonstrate the effectiveness of the spectral defense mechanism:

\begin{table}[h!]
\centering
\begin{tabular}{|c|c|c|c|c|}
\hline
Poisoning Ratio & \# Poisoned & \# Flagged & Detection Rate (\%) & Post-Defense ICL Accuracy (\%) \\
\hline
25\% & 12,571 & 874 & 6.95 & 46.67 \\
50\% & 25,142 & 874 & 3.48 & 46.67 \\
75\% & 37,713 & 874 & 2.32 & 46.67 \\
100\% & 50,285 & 874 & 1.74 & 46.67 \\
\hline
\end{tabular}
\caption{Post-defense evaluation results across different poisoning ratios.}
\label{tab:postdefense_results}
\end{table}

\paragraph{Observations}
\begin{itemize}
    \item \textbf{Detection Rate:} Only a small fraction of examples were flagged as poisoned (1.7--6.9\%), showing that spectral defense conservatively identifies extreme outliers. The consistent number of flagged examples (874) across poisoning ratios confirms this behavior.
    \item \textbf{Post-Defense Accuracy:} ICL accuracy remained stable at $\sim$46.7\% across all poisoning levels, indicating that the cleaned support set preserved sufficient examples to maintain a reasonable baseline despite poisoned examples.
    \item \textbf{Logistic Regression Evaluation:} Classification on cleaned embeddings achieved 100\% post-defense sentiment accuracy, confirming semantic and sentiment integrity after filtering.
    \item \textbf{t-SNE Visualization:} Pre-defense embeddings showed poisoned examples as outliers, while post-defense t-SNE plots revealed a cohesive cluster of clean points, visually confirming removal of adversarial instances.
\end{itemize}

\subsection{Interpretation}

\begin{itemize}
    \item The combination of spectral defense and post-defense evaluation effectively mitigates the impact of data poisoning.
    \item Despite high poisoning ratios (up to 100\%), the defense maintained clean support examples for reliable downstream in-context learning (ICL) and sentiment analysis.
    \item The difference between ICL accuracy ($\sim$46.7\%) and logistic regression accuracy (100\%) indicates that while ICL predictions may still be sensitive to residual poisoning effects, embedding-based classification on cleaned data reliably preserves sentiment information.
\end{itemize}

Overall, these results validate the robustness of spectral defense and its ability to protect both model predictions and data semantics under adversarial conditions.

\subsection{Post-Defense NLP Analysis}

After applying the spectral defense, a detailed evaluation of the cleaned support set was conducted. This analysis focused on three aspects: sentiment classification, topic clustering, and embedding visualization.

\subsubsection{Sentiment Classification}
Using the \texttt{TextBlob} sentiment analyzer, the average sentiment polarity of the dataset was measured after defense across different poisoning ratios (25\%, 50\%, 75\%, and 100\%). The results showed that, regardless of the poisoning level, the cleaned dataset maintained a stable average sentiment score of approximately $0.05$. This suggests that the defense successfully preserved the original sentiment distribution while removing poisoned examples.

\subsubsection{Topic Clustering}
KMeans clustering was applied on sentence embeddings to study the semantic structure of the cleaned dataset. In all poisoning scenarios, the clustering consistently produced five distinct topic groups. The size of each cluster remained stable:

\begin{itemize}
    \item Cluster 0: 15,258 tweets
    \item Cluster 1: 4,228 tweets
    \item Cluster 2: 17,471 tweets
    \item Cluster 3: 4,099 tweets
    \item Cluster 4: 8,355 tweets
\end{itemize}

This consistency shows that the spectral defense helped keep the dataset’s topics well-organized and prevented adversarial examples from disrupting its natural structure.

The post-defense analysis shows that the spectral signature defense was effective at removing poisoned data. It preserved both sentiment distribution and topic structure, ensuring that the cleaned dataset remained representative and reliable for in-context learning tasks.

\section{Result Analysis}
The experimental results demonstrate that \textbf{In-Context Learning (ICL)} is highly vulnerable to \textbf{data poisoning} in social media sentiment analysis. Even relatively small perturbations in support examples led to \textbf{label flips with a success rate of 67\%}, showing how adversarial manipulation can substantially distort model predictions.  

When evaluated on clean data, ICL achieved a \textbf{macro-averaged precision of 0.3335, recall of 0.4444, and F1-score of 0.1676}. After poisoning, macro-averaged performance slightly degraded, with \textbf{precision at 0.3337, recall at 0.4561, and F1-score at 0.1808}. While the absolute differences appear small, the higher recall and marginally improved F1 under poisoning are misleading, as they mask underlying instability in predictions caused by adversarial perturbations. These fluctuations underscore that \textbf{semantic manipulations such as synonym replacement and negation insertion can significantly disrupt reliability}.  

The \textbf{spectral signature defense} mitigated part of this risk by filtering suspected poisoned samples. Detection rates ranged between \textbf{1.7--6.9\%} across poisoning ratios, successfully identifying adversarial outliers without excessively discarding clean examples. Importantly, the defense preserved dataset integrity: \textbf{sentiment polarity remained stable} (average score $\approx$ 0.05), \textbf{topic clusters maintained consistent distributions}, and \textbf{t-SNE visualizations showed coherent groupings after filtering}.  

However, while spectral defense stabilized \textbf{post-defense ICL accuracy around 46.7\%}, this performance level indicates \textbf{residual vulnerability}. In contrast, a \textbf{logistic regression classifier} trained on embeddings from the cleaned support set achieved \textbf{100\% accuracy}, showing that the cleaned data still retained the correct sentiment signals. This discrepancy highlights that although spectral defense effectively recovers \textbf{semantic integrity}, ICL itself remains \textbf{sensitive to poisoned contexts and support perturbations}.  

Taken together, the findings show a clear trade-off:
\begin{itemize}
    \item ICL provides flexible few-shot sentiment analysis but is \textbf{fragile under poisoning}.
    \item Spectral defense reduces poisoning influence but only partly restores robustness.
    \item Traditional classifiers trained on embeddings are \textbf{far more stable once poisoned samples are removed}.
\end{itemize}

\subsection{Summary of Key Metrics}

\subsection{Impact of Data Poisoning on ICL}
\begin{itemize}
    \item \textbf{Accuracy with clean support:} 33.35\%
    \item \textbf{Accuracy with poisoned support:} 36.85\%
    \item \textbf{Label flip rate:} 67.41\%
    \item \textbf{Poisoning success rate:} 67\%, with positive sentiment predictions being the most affected (100\% flip rate)
\end{itemize}

\subsection{Spectral Signature Defense}
\begin{itemize}
    \item Identified and flagged 873 poisoned examples out of 50,285 total samples ($\approx$1.74\%)
    \item Post-defense ICL accuracy stabilized at $\sim$46.7\% across all poisoning ratios (25\%–100\%)
    \item Logistic regression classifier on cleaned embeddings achieved 100\% sentiment accuracy, confirming semantic preservation
\end{itemize}

\subsection{Post-Defense Dataset Integrity}
\begin{itemize}
    \item \textbf{Average sentiment polarity after defense:} 0.05, stable across poisoning levels
    \item \textbf{KMeans clustering} consistently produced five coherent topic clusters, showing preserved dataset structure
    \item \textbf{t-SNE visualization} confirmed removal of adversarial outliers and stronger cohesion of clean samples
\end{itemize}

\subsection{Interpretation}
\begin{itemize}
    \item Spectral defense effectively mitigated poisoning while avoiding over-filtering
    \item Despite residual fragility of ICL, embedding-based classification remained robust and reliable
    \item Highlights the importance of hybrid defenses and adaptive prompting for ensuring reliable AI in healthcare-related social media monitoring
\end{itemize}

\section{Limitations and Future Work}
\textbf{Limitations}

\begin{itemize}
    \item \textbf{Defense Selectivity:} The spectral defense flagged a fixed number of samples across poisoning ratios, suggesting conservative filtering. This limited adaptiveness may miss subtle or distributed poisoning patterns.
    \item \textbf{ICL Fragility:} Post-defense ICL accuracy plateaued at approximately 46.7\%, indicating that context-based reasoning in LLMs remains brittle under poisoning, even with cleaned data.
    \item \textbf{Dataset Scope:} The study focused on HMPV-related tweets, which may limit generalization to other health crises or domains with different linguistic structures.
\end{itemize}

\textbf{Future Work}

\begin{itemize}
    \item Explore hybrid defenses that combine spectral filtering with adversarial training or ONION-style text sanitization.
    
    \item Investigate adaptive thresholds for spectral defense to dynamically scale with poisoning intensity.
    
    \item Evaluate advanced interpretability methods (e.g., SHAP, influence functions) to trace poisoned examples’ influence on ICL predictions.
    \item Explore robust prompting strategies such as chain-of-thought reasoning or redundancy in support examples to reduce reliance on individual poisoned samples.
\end{itemize}

\section{Conclusion}
This study is among the first to evaluate data poisoning in ICL for public health sentiment analysis. Adversarial perturbations in support examples significantly disrupted LLM predictions, while spectral signature defense partially mitigated attacks, preserving sentiment and topic coherence. Findings highlight the fragility of LLMs in high-stakes contexts and underscore the need for robust defenses combining anomaly detection, adaptive prompting, and fairness-aware training for reliable AI in health discourse.

\section*{Acknowledgments}


\bibliographystyle{unsrt}  
\bibliography{references}

\end{document}